\documentclass{article}
\usepackage{ijcai18}

\usepackage{times}
\usepackage{xcolor}
\usepackage{soul}
\usepackage[utf8]{inputenc}
\usepackage[small]{caption}
\usepackage{epsfig}
\usepackage{graphicx}
\usepackage{subfig}
\usepackage{float}
\usepackage{makecell}
\usepackage{ booktabs}
\usepackage{amsmath}
\usepackage{amssymb}

\title{Out of the Black Box: Properties of deep neural networks and their applications}

\author{
Nizar Ouarti$^1$, 
David Carmona$^1$, 
\\ 
$^1$ IPAL, Sorbonne Universite, CNRS, NUS \\
nizar.ouarti@ipal.cnrs.fr,
david.carmona93@gmail.com
}
\begin{document}

\maketitle

\begin{abstract}  
   Deep neural networks are powerful machine learning approaches that have exhibited excellent results on many classification tasks. However, they are considered as black boxes and some of their properties remain to be formalized. In the context of image recognition, it is still an arduous task to understand why an image is recognized or not. In this study, we formalize some properties shared by eight state-of-the-art deep neural networks in order to grasp the principles allowing a given deep neural network to classify an image. Our results, tested on these eight networks, show that an image can be sub-divided into several regions (patches) responding at different degrees of probability (local property). With the same patch, some locations in the image can answer two (or three) orders of magnitude higher than other locations (spatial property). Some locations are activators and others inhibitors (activation-inhibition property). The repetition of the same patch can increase (or decrease) the probability of recognition of an object (cumulative property). Furthermore, we propose a new approach called Deepception that exploits these properties to deceive a deep neural network. We obtain for the VGG-VDD-19 neural network a fooling ratio of 88\%. Thanks to our "Psychophysics" approach, no prior knowledge on the networks architectures is required.
  
\end{abstract}

\section{Introduction}

A Multilayer Perceptron is a multidimensional universal function approximator  \cite{cybenko1989approximation} that learns a mapping between an input and an output. 
However, the black box problem stresses that it is ordinarily a challenge to understand what the network exactly learns.  
One advantage of classical handcrafted approaches is to be able to guarantee a perfectly predictable behavior. This is critical for some areas like medicine or aviation. 

Deep Neural Networks (DNN) inherit some issues of Multilayer Perceptron and particularly the black box problem. However, is it possible to obtain a more explainable AI based on DNN? Our motivation is to propose some methods and properties that help users and designers to understand more formally the behavior of DNNs. 

In this article our main contributions are the followings:
    \begin{itemize}
    	\item We propose a new methodology that is the Psychophysics on Deep Neural Networks.
        \item We formalize and quantify some important properties of the Deep Neural Networks.
        \item We show an extensive experimental comparison illustrating that the mentioned properties are followed by eight state-of-the-art deep neural networks.
        \item We propose a new strategy, called Deepception, based on the formalized properties and capable of fooling a deep neural network, even if its architecture is unknown.
    \end{itemize}

    \begin{figure}[t]
        \centering
        \includegraphics[width=1\linewidth]{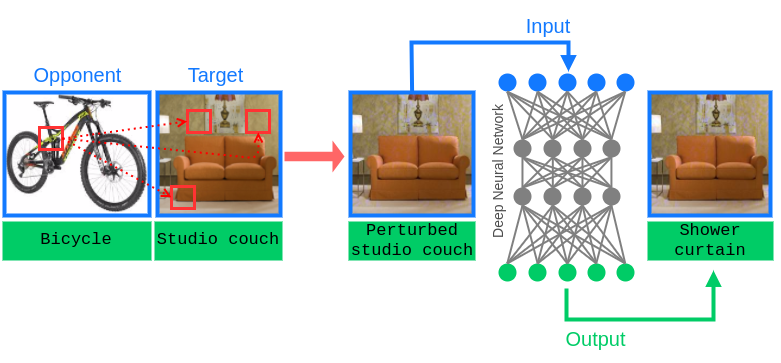}
        \caption{Depiction of the Deepception's pipeline. It uses all the four properties reported in this paper: the local, spatial, cumulative and activation-inhibition properties. We extract a sub-part of the opponent image called the patch. Then, we insert the patch into the target at positions where it is imperceptible to human eyes. The resulting perturbed image is misclassified by the Deep Neural Network.}
        \label{fig:first_Page}
    \end{figure}

\section{Related works}
Deep neural networks exhibit the best performances for image classification with different datasets like ImageNet\cite{deng2009imagenet,ILSVRC15,krizhevsky2012imagenet,simonyan2014very,he2016deep} or Pascal VOC for recognition and detection \cite{ren2015faster}. However, one of the principal limitations is the lack of understanding of what the networks are really learning. This is usually referred as the black box problem. This problem leads to some limitations.
One is related to the reliability of the probability. As pointed out by Gal and Gharamani (2016) \cite{gal2016dropout} the standard deep learning tools designed for regression and classification do not effectively capture model uncertainty. They argue that the probability, that is the output of the network, cannot be considered as the confidence of the model. However, for some critical software like autopilot or medical software, it is a requirement to have an accurate estimation of this uncertainty. 

Another limitation is related to predictability. It is indeed very difficult to understand why some objects are recognized and others are not. To illustrate this issue, Nguyen et al. (2015) \cite{nguyen2015deep} show that some abstract geometric patterns could lead the network to recognize an object not present in the image. They also show that a periodic pattern is sufficient to mislead a deep neural network.

The fooling is also a critical issue. It is shown that with a small magnitude adversarial noise, it is possible to change the ranking of the classes \cite{moosavi2016deepfool,moosavi2016universal,lin2017tactics}. Even a random noise can provoke the same effect \cite{fawzi2016robustness}.

In the classical scientific approach, scientists are able to quantify and understand some parameters or properties that have an important impact on the results. Deep neural networks are more complex and the non-linearity of the processes seems to prevent any thorough analysis.  

Some studies address this question of black box by visualizing the filters of the networks \cite{zeiler2014visualizing}. Other ones visualize and analyze the different layers of the network, to discover the pixel that have a more relevant impact on the classification \cite{yosinski2015understanding,binder2016analyzing,nguyen2016multifaceted,samek2017evaluating}. These approaches are not using patches and can be considered as neurophysiological approaches for DNN in our formalism.
This paper uses a totally different approach based on methodological reductionism. The goal is to identify some fundamental properties of deep neural networks. This novel approach can be compared to Psychophysics \cite{fechner1860elemente,stevens1960psychophysics,henn1975ernst}. The observation of thresholds of perception related to physical stimuli is replaced by the monitoring of the probability score of a class related to the manipulation of some visual inputs. One consequence is that no prior knowledge about the architecture of the network is required and the method can be applied to any deep neural network.
%
%
\section{Our approach}
In this paper, we formalize some properties that have an important impact on DNNs: 

    \begin{itemize}
        \item An image can be sub-divided into a group of regions responding at different levels of probability. The size of the patch is related to the probability. We called it: the Local property.

        \item  When an identical patch is positioned at different locations of an image, its probability varies. We called it: the Spatial property.
        
        \item The repetition of the same patch can increase (or decrease) the probability of recognition of an object. This repetition can be added at different location of the image. We called it: the Cumulative property.
        
        \item  Cumulative property can have an activator or inhibitor behavior related to the location of the next patch. This location of activation and inhibition is not related to the initial spatial probability with one patch. We called it: the Activation-Inhibition property.
        
    \end{itemize}

\section{Methodology}
The psychophysical-driven approach proposed in this paper highlights the existence of the four properties. Our strategy is to modify a physical aspect of the image inserted into the deep neural network and observe how its probability varies accordingly to the modification we made. The main concern of this study is the relationship between the network's input and its output. Consequently, the approach presented in this paper does not require a prior knowledge of the targeted deep neural network architecture.

    \subsection{Deep neural networks models}
    
        In order to investigate the generalization of the four properties, we decide to run the experiments on eight state-of-the-art deep neural networks applied to Pascal VOC 2007 and ILSVRC 2012 ImageNet datasets. 
        
        We consider the following architectures on \textbf{Pascal VOC 2007} three MatConvnet pre-trained networks: Caffenet, VGG16 and VGGM12\footnote{Their mean Average Precisions are 57.3\%, 67.3\% and 59.4\% respectively.}. We refer to this ensemble of models as "Fast-RCNN networks". On \textbf{ILSVRC 2012 ImageNet},  we consider five MatConvnet pre-trained networks: GoogleLeNet, ResNet152, Vgg-f, Vgg-Verydeep-16 and Vgg-Verydeep-19\footnote{Their top-1 error rates are 34.2\%, 23\%, 41.1\%, 28.5\% and 28.7\% respectively.}. We call this set of models "ImageNet networks" throughout the paper.
    \subsection{Object categories}
            
        We choose from Internet the images of objects used in the experiments. They are all grayscale images. This paper is dealing with two different architectures of deep neural networks: Pascal VOC 2007 and ILSVRC 2012 ImageNet. Consequently, it is important to take objects belonging to two distinct sets of categories. 
        The categories chosen for Pascal are: \textit{bird, person, cat, cow, dog, horse, sheep, airplane, bicycle, bus, car, motorbike, train, bottle, chair, dining table, potted plant, sofa, tv monitor, boat.}     
        For ImageNet, we pick a subset of the 1000 categories of ImageNet. These 20 categories are displayed in the Figure  \ref{fig:patches}.
            
        We take a subset because processing the entire set would be computationally expensive. However, we respect the same object-type proportions than the Pascal categories: eight categories correspond to biological objects (i.e. bison, daisy or bald eagle) and twelve are manufactured (i.e. airliner, bullet train or cab).
        The patches are extracted with the following method. For the 8 deep neural networks and the 20 selected categories, we choose four different grayscale images. To minimize the effects related to the size of the images, we start by resizing the input images 600x600 pixels if the DNN is a Fast-RCNN network and 224x224 pixels if it is an ImageNet networks. We create several sliding cropping windows which sizes depend on the tested DNN and do not overlap each other: 50x50,100x100,150x150, 200x200 pixels If the DNN is a Fast-RCNN network and 37x37, 56x56, 75x75, 112x112 pixels if the DNN is an ImageNet one. The deep neural network assigns a probability and a label each cropped patch. Then, the algorithm selects among all the patches extracted from the four images, the one having the best probability for the category of the object. In the end of the process, we will have in total 20 patches for each DNN. Figure \ref{fig:patches} gives an example of extracted patches for the VGG-VDD-19 network and their probability.
%
%
\section{Results}
        
    \subsection{Local property}

\begin{figure}[t]
            \centering
            \includegraphics[width=0.8\linewidth,height=3.5cm]{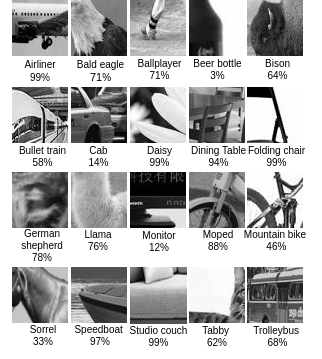}
            \caption{Patches extracted by our system using the VGG-VDD-19 network. The labels underneath correspond to the classes they belong to. Most of them are not directly recognizable by a human.}
            \label{fig:patches}
        \end{figure}

\begin{figure}
            \centering
            \subfloat[Fast-RCNN CNNs]{%
                \includegraphics[width=0.5\linewidth]{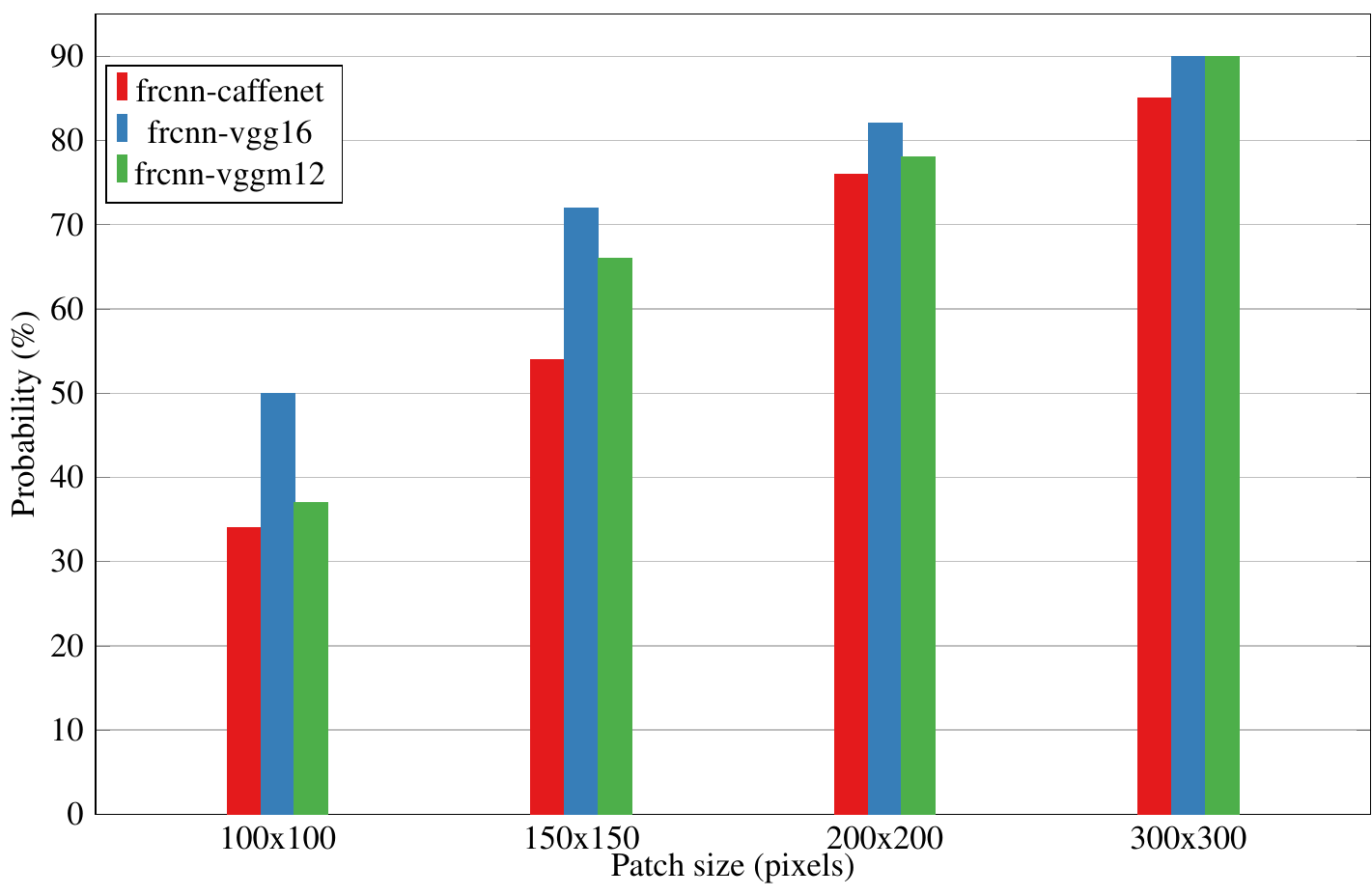}%
                \label{fig:pascal_scale}%
            }\hfill
            \subfloat[ImageNet CNNs]{%
              \includegraphics[width=0.5\linewidth]{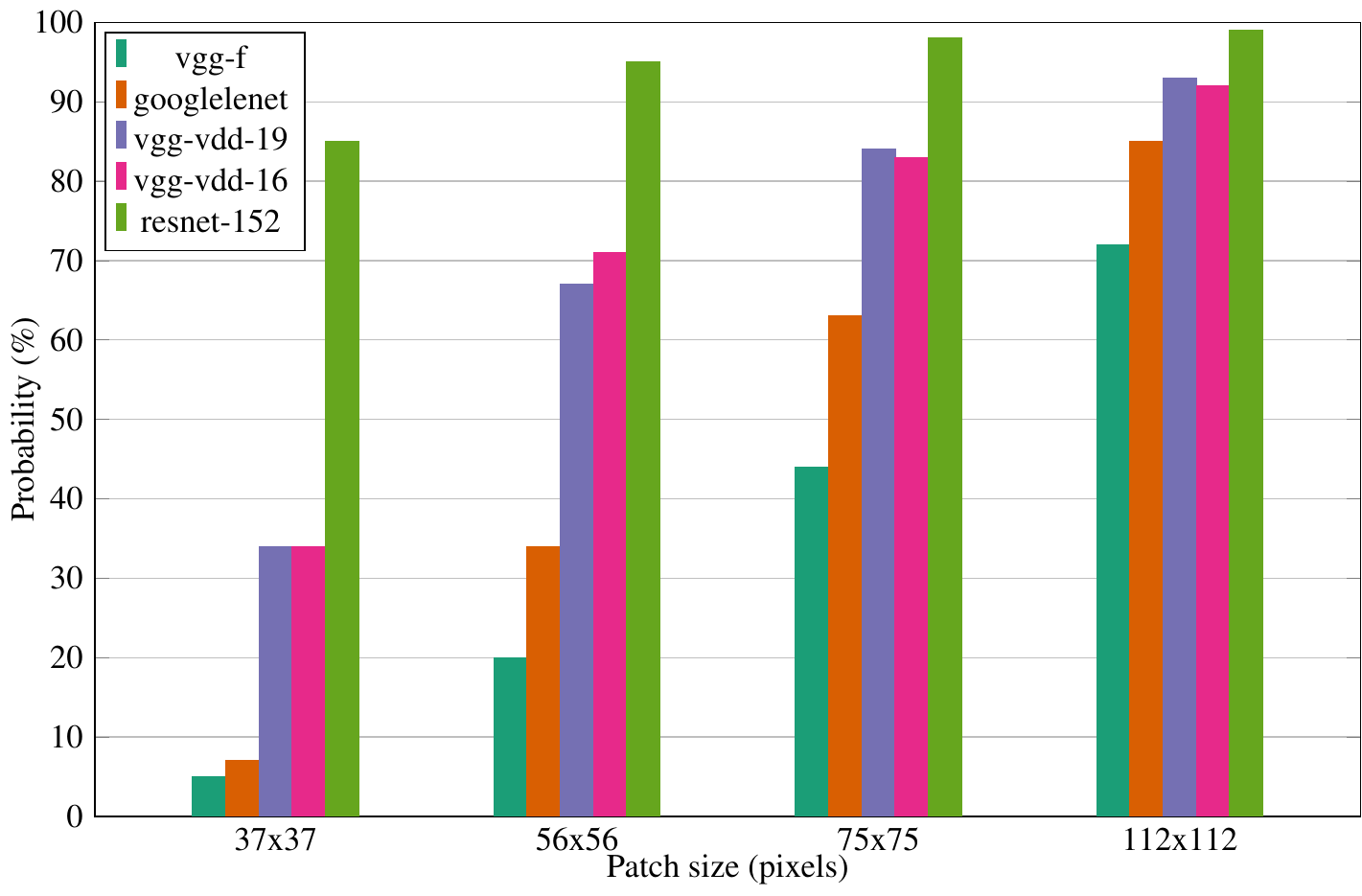}%
              \label{fig:imagenet_scale}%
            }
            \caption{Average probabilities of the patches for each deep neural network for different scales.}
            \label{fig:scale}
        \end{figure}

        A patch is a sub-part of the image. It has two principal characteristics: a size and a probability belonging to a specific category. 
        Different patches have different probabilities. This is illustrated in Figure \ref{fig:patches} were we display after an exhaustive research the patches that answered with the highest probability.  A second observation is that the probability for a given patch is not scale invariant (see Figure \ref{fig:scale}). Moreover this probability is very different if the patch is resized before being sent to the network (Figure \ref{fig:scale})  with a range between 5 and 99 percent or if the patch is not resized and sent with all the other pixel put at zero (Figure \ref{fig:add_pixel_pixel_surf}) with a range between 0.6 and 1.5 $10^{-4}$. 

        Figure \ref{fig:scale} shows how the average probability of the patches vary according to their sizes. Bigger patches have higher probabilities, however it is still possible with a relatively small patch to obtain an important probability. 
        For the rest of the experiments, we decided to take patches of size 150x150 and 56x56 pixels for the Fast-RCNN networks and ImageNet respectively. We chose these sizes because they are challenging to recognize by humans, and at the same time, correctly detected by the DNNs.
        
    \subsection{Spatial property}
\begin{figure}[H]
            \centering
            \includegraphics[width=0.7\linewidth,height=3.5cm]{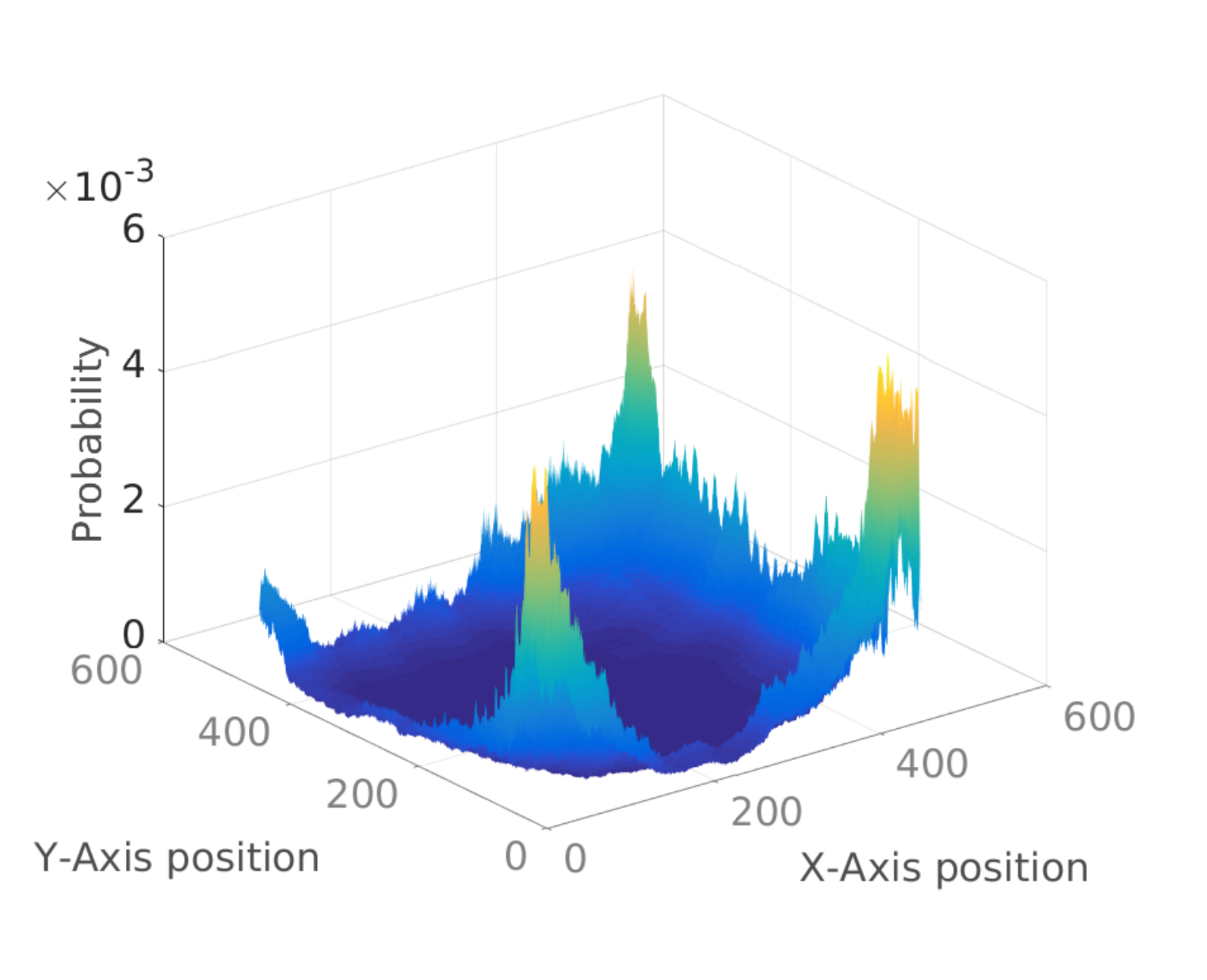}
            \caption{Evolution of the probability according to a (150X150) patch position on the 600x600 black image. The high ratio ($Proba_{max}/Proba_{min}$),3769,indicates that Fast-RCNN-VGG16 is sensitive to the spatial property.}
            \label{fig:add_pixel_pixel_surf}
        \end{figure}

        To demonstrate whether DNNs answer differently for different location of the patch inside an image, we create a black image (i.e. all the pixels have values equal to zero) of size 600x600 pixels (if we use a Faster-RCNN network) or 224x224 pixels (if we are using an ImageNet network). 
        Then, we add in a pixel-wise fashion the patch by the corresponding region of the black image where it should be positioned. We iterate until we reach the end of the image.   
        We decided to perform on one image a dense mapping to observe the effect of location on the probability. The patch is a 150X150 patch of a cat detected initially at 99\% by the Fast-RCNN-VGG16 network. This procedure takes a long time and cannot be performed on many images. Figure \ref{fig:add_pixel_pixel_surf} gives an interesting evaluation of how extreme the differences of probability can be. With the same patch, the range of probability was from 0.6 and 1.5 $10^{-4}$. The ratio between the maximum and minimum is 3769 (i.e. the maximum probability is 3769 times higher than the minimum probability).

        To be more exhaustive and to test more networks, we decided to run this experiment on many images and with many patches but less densely (no overlap of the patch), i.e. 16 positions for a 150X150 patch on a 600X600 image. The max-min ratio for different networks can be observed in Table \ref{tab:spatial_table}.

 For some categories, deep neural networks are very sensitive to the spatial property. Notice that a state-of-the-art neural network as ResNet-152 can be extremely sensitive to the spatial property. For the 'moped' category, the maximum probability is 10000 higher than the minimum.

        Using this procedure, we observe that all the ratios between the region that maximize and minimize the patch probability are always higher than 1. 
        
        We report in Table \ref{tab:spatial_table} the average ratios for all the deep neural networks. We can notice that there are different degrees of sensitivity to the spatial property. However, the most sensitive DNN to the spatial property is ResNet-152. In that case, the maximum probability is 2262 times higher than the lowest.
        
        \begin{table}[t]
            \begin{center}
            \begin{tabular}{c c c c}
            \Xhline{2.5\arrayrulewidth}
            {\scriptsize Deep neural network} & {\scriptsize Avg. Ratio} & {\scriptsize Avg. Maximum} & {\scriptsize Avg. Minimum} \\
            \hline
            {\scriptsize Fast-RCNN-CaffeNet} & {\scriptsize156.98} & {\scriptsize 0.0047} & {\scriptsize 0.00003}\\
            {\scriptsize Fast-RCNN-VGG16} & {\scriptsize 572.35} & {\scriptsize 0.0007} & {\scriptsize 0.000001}\\
            {\scriptsize Fast-RCNN-VGGM12} & {\scriptsize 1237.35} & {\scriptsize 0.0005} & {\scriptsize 0.0000004}\\
            {\scriptsize Vgg-f} & {\scriptsize 20.15} & {\scriptsize 0.0003} & {\scriptsize 0.00002}\\
            {\scriptsize GoogleLeNet} & {\scriptsize 60.11} & {\scriptsize 0.0017} & {\scriptsize 0.00003}\\
            {\scriptsize Vgg-VDD-19} & {\scriptsize 208.54} & {\scriptsize 0.0023} &{\scriptsize 0.00001}\\
            {\scriptsize Vgg-VDD-16} & {\scriptsize 161.57} & {\scriptsize 0.0034} &{\scriptsize 0.00002}\\
            {\scriptsize ResNet-152} & {\scriptsize 2262.53} & {\scriptsize 0.0115} &{\scriptsize 0.000005}\\
            \Xhline{2.5\arrayrulewidth}
            \end{tabular}
            \end{center}
            \caption{Averages of the ratios and gains for the spatial, activation and inhibition experiments.}
            \label{tab:spatial_table}
        \end{table}

    \subsection{Cumulative Property and Activation-Inhibition Property}
        In this section, the aim is to show that the addition of the same patch many times will change the probability of detection. We decided to work with a black image of size 600x600 or 224x224 pixels accordingly to the used deep neural network. Then, the image is divided into 16 non-overlapping areas of dimensions 150x150 or 56x56 pixels each. We begin by placing the patch inside the area that maximizes the probability. Afterwards, we look for an additional patch position that increases the probability and add the patch again. We keep performing this operation while the probability is increasing after a new placement. 
        Because we wanted to show the Activation-Inhibition effect, we apply the same algorithm for inhibition. Obviously, instead of looking for the areas that increase the overall probability, we search for the areas making the overall probability decrease.

\begin{figure*}
        \centering
        \subfloat[Activation Fast-RCNN]{%
            \includegraphics[width=0.24\linewidth,height=4.5cm]{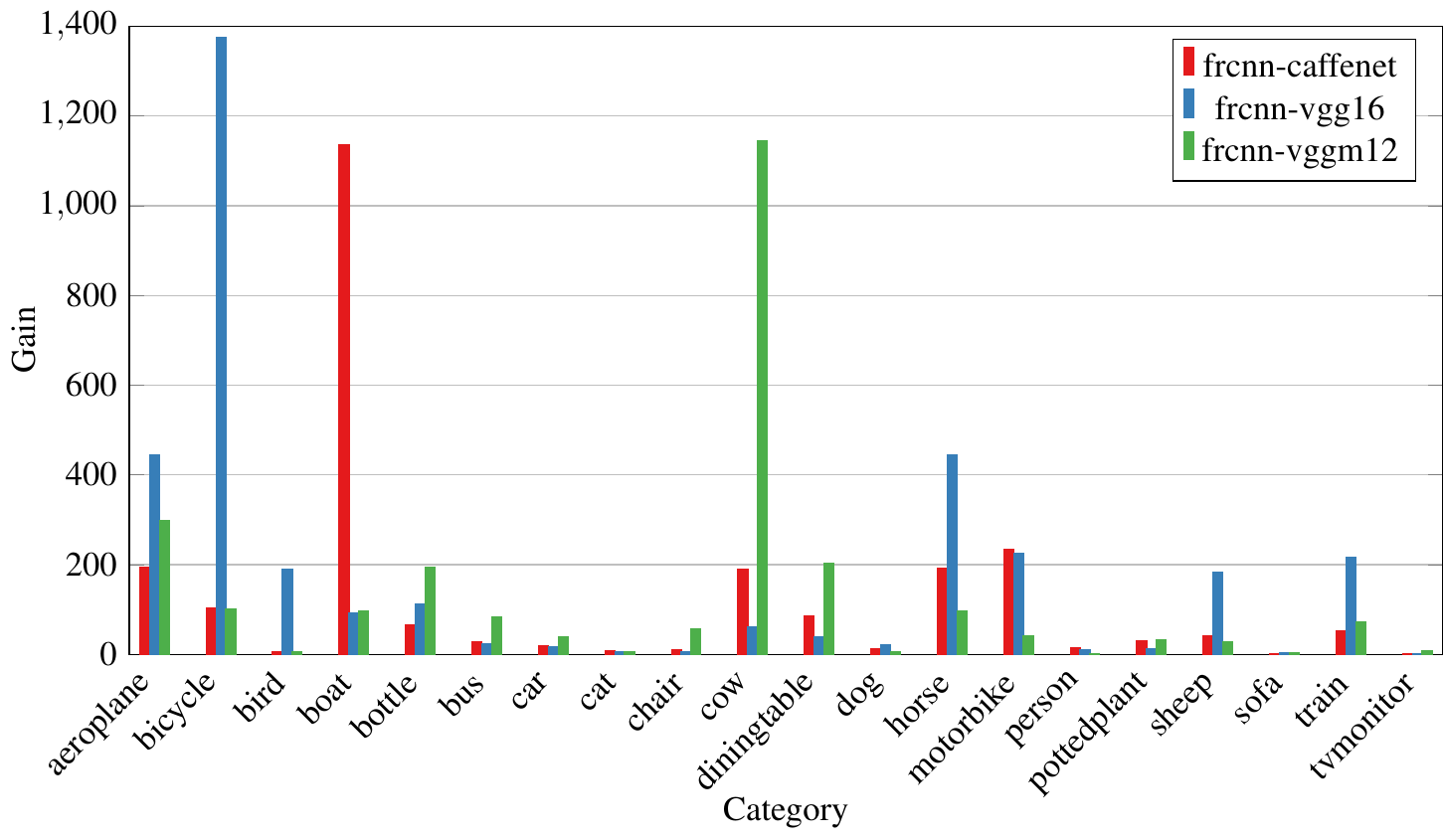}%
            \label{fig:gains_activations_fast}%
        }
        \subfloat[Inhibition Fast-RCNN]{%
          \includegraphics[width=0.24\linewidth,height=4.5cm]{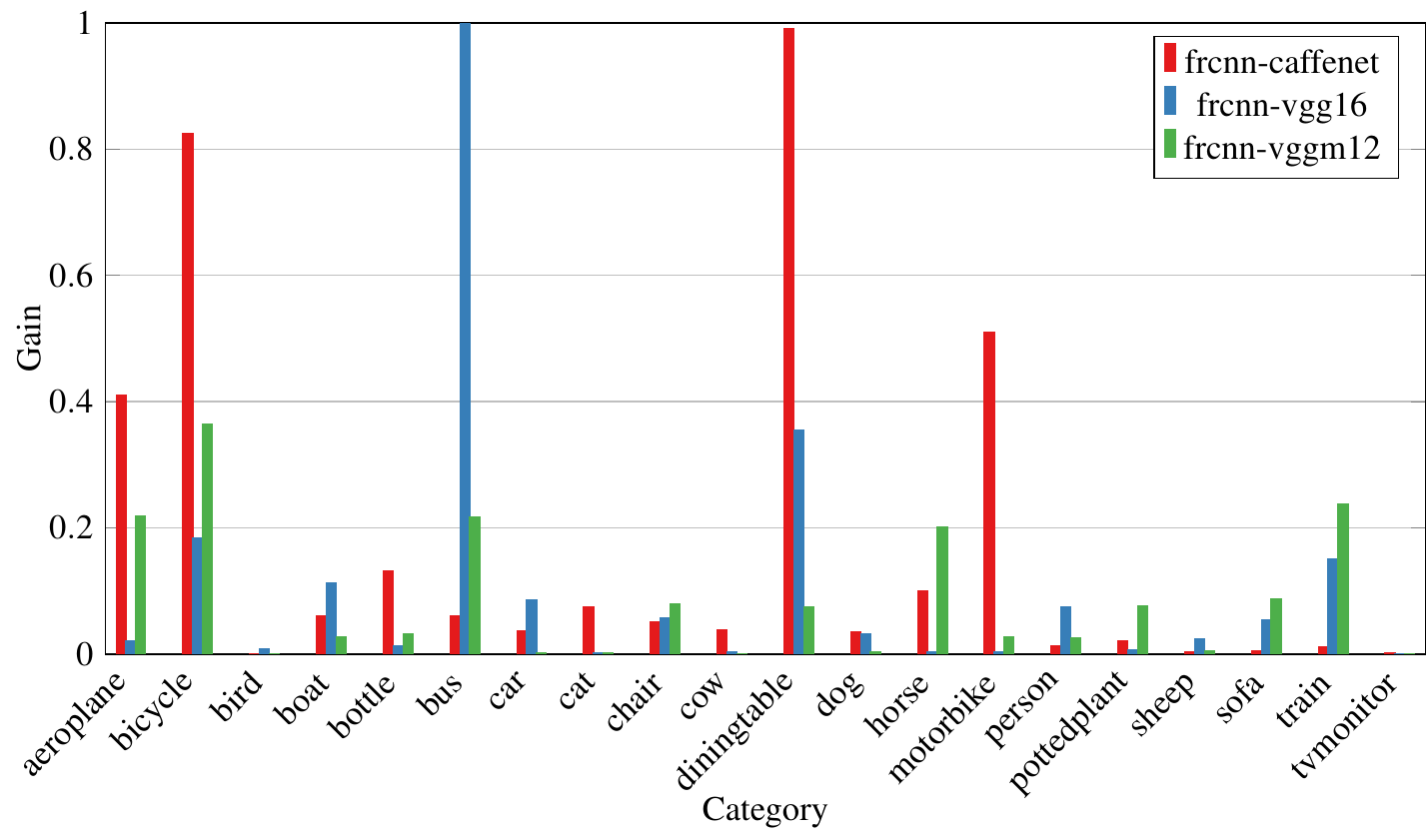}%
          \label{fig:gains_inhibition_fast}%
        }
        \subfloat[Activation ImageNet]{%
          \includegraphics[width=0.24\linewidth,height=4.5cm]{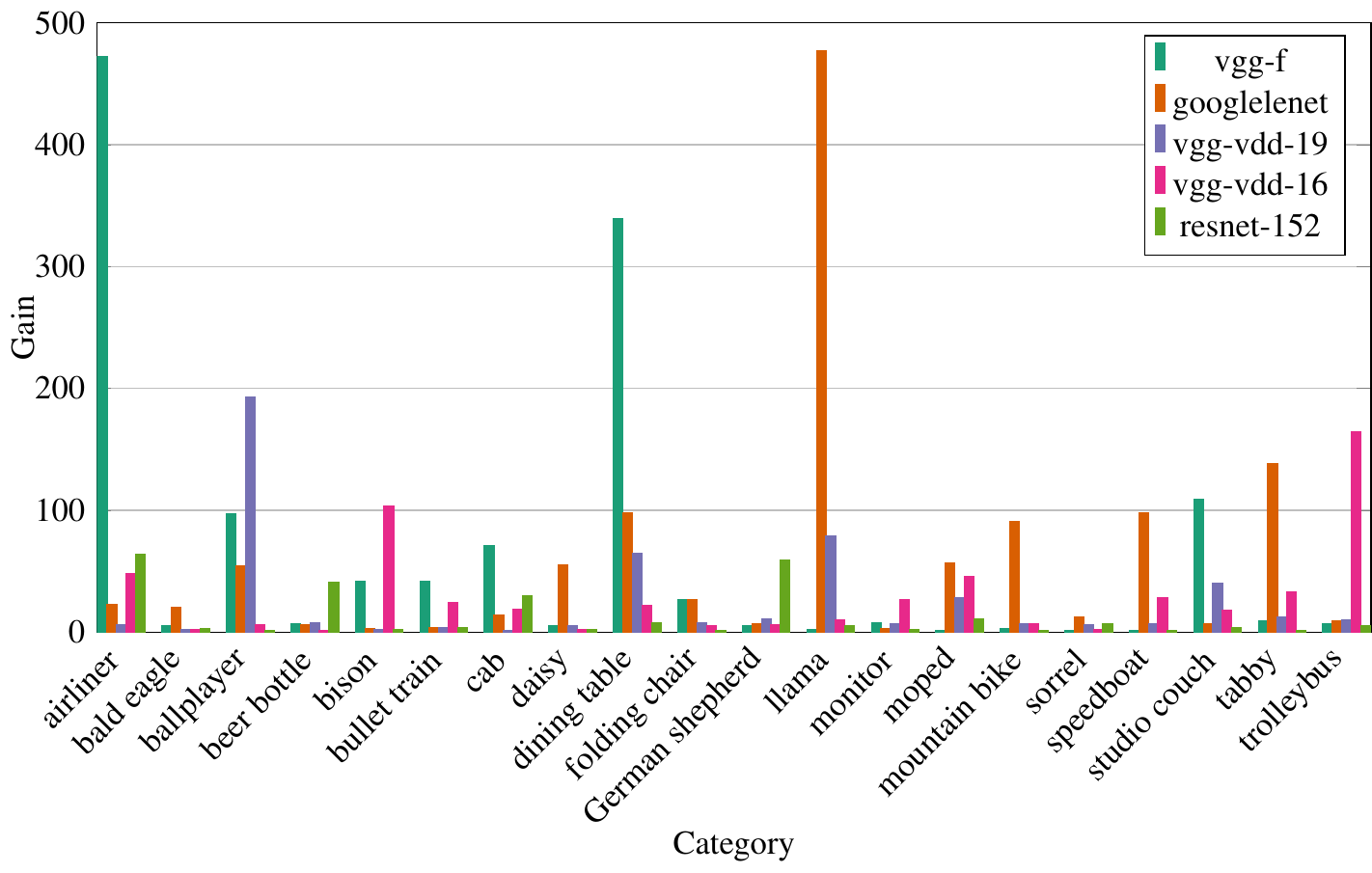}%
          \label{fig:gains_activations_imagenet}%
        }
        \subfloat[Inhibition ImageNet]{%
          \includegraphics[width=0.24\linewidth,height=4.5cm]{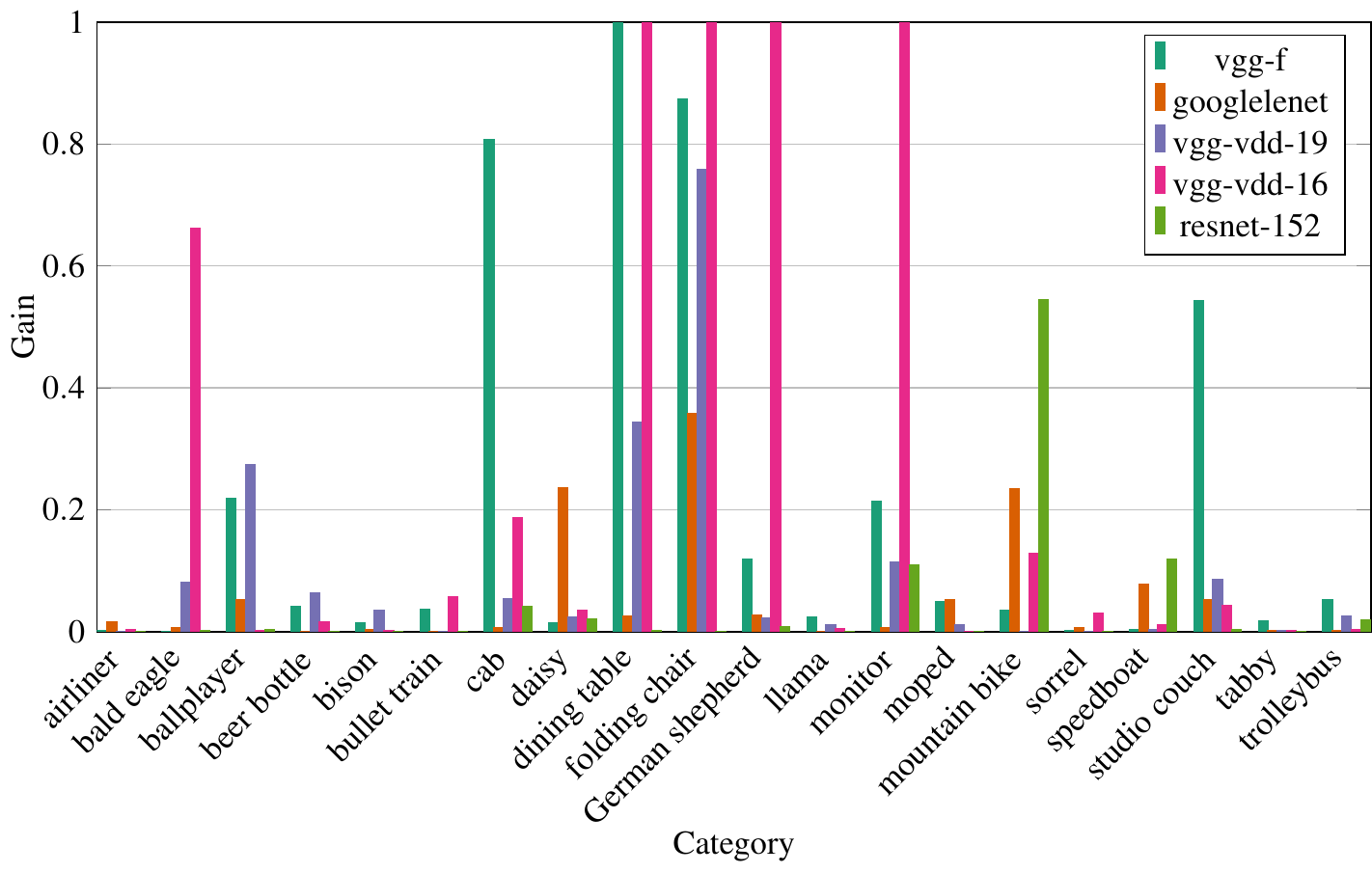}%
          \label{fig:gains_inhibitions_imagenet}%
        }
        \caption{Gains for the activation and inhibition experiments. All the gains are higher than one in the case of activation and lower than one in the case of inhibition. Consequently, it means that all the deep neural networks follow both properties independently of the patch.}
        \label{fig:activation_inhibition}
    \end{figure*}

        The gain is defined as the relation between the first best positioning and the final probability: $Gain.Prob_{init}=Prob_{final} $. It is superior to 1 for activation and inferior to 1 for inhibition. Figure \ref{fig:activation_inhibition} reports the gains for all the tested categories and DNNs. For both types of neural networks, the gains are always higher than 1. This means that the final probability will always be increased after a sequence of patch placement compared to one single position.
        Furthermore, it is also possible to decrease the probability of the class. Figure \ref{fig:activation_inhibition} proves the probability of placing multiple patches can decrease the probability. We show with these experiments that it is possible to change the probability of recognition by adding new patches (cumulative property). But we also show that an oriented strategy of placement of the patch can increase or decrease the probability (activation-inhibition property).

    \subsection{Deceiving a deep neural network: Deepception}
    
    To show a practical usage of the properties we highlighted, we decided to design a new type of fooling algorithm called Deepception. This algorithm has the particularity to be independent from the architecture, thanks to our psychophysic approach. We need simply an access to the input (image) and the output of the network (probability of recognition) to deceive the DNN. It means that our approach can even deceive an unknown DNN that is on a server. The idea is that we can pick some local patches (local property) that have by themselves a high probability of recognition for a given class. And we insert these patches in a targeted image (that belongs to another class) to fool the DNN (see Figure \ref{fig:first_Page}). We design a specific cost function (equation \ref{eq:cost}) that encourages the fooling by estimating the probability for different spatial insertions. Here, we take advantage of the cumulative and activation-inhibition property. The inserted normalized patches are made transparent to not be perceptible by humans. In order to do so, the patch is multiplied by transparency coefficient $\tau$. This resulting patch, called the decoy, is multiplied with the weaker RGB channel.
 The cost function is: 
 \begin{equation}
 \underset{N,L \in \mathbb{N}}{\mathrm{Argmin}}  (P_{t} ), \qquad stopping \quad criterion \quad P_{t}>P_{i}
 \label{eq:cost}
  \end{equation}
Where N represents the number of patches and L the location of the patches, $P_{t}$  represents the probability of the targeted class. This targeted class is the initial class of the image. $P_{i}$ is the probability of each of the other classes.

\begin{figure}[H]
                \centering
                \includegraphics[width=0.8\linewidth,height=4.0cm]{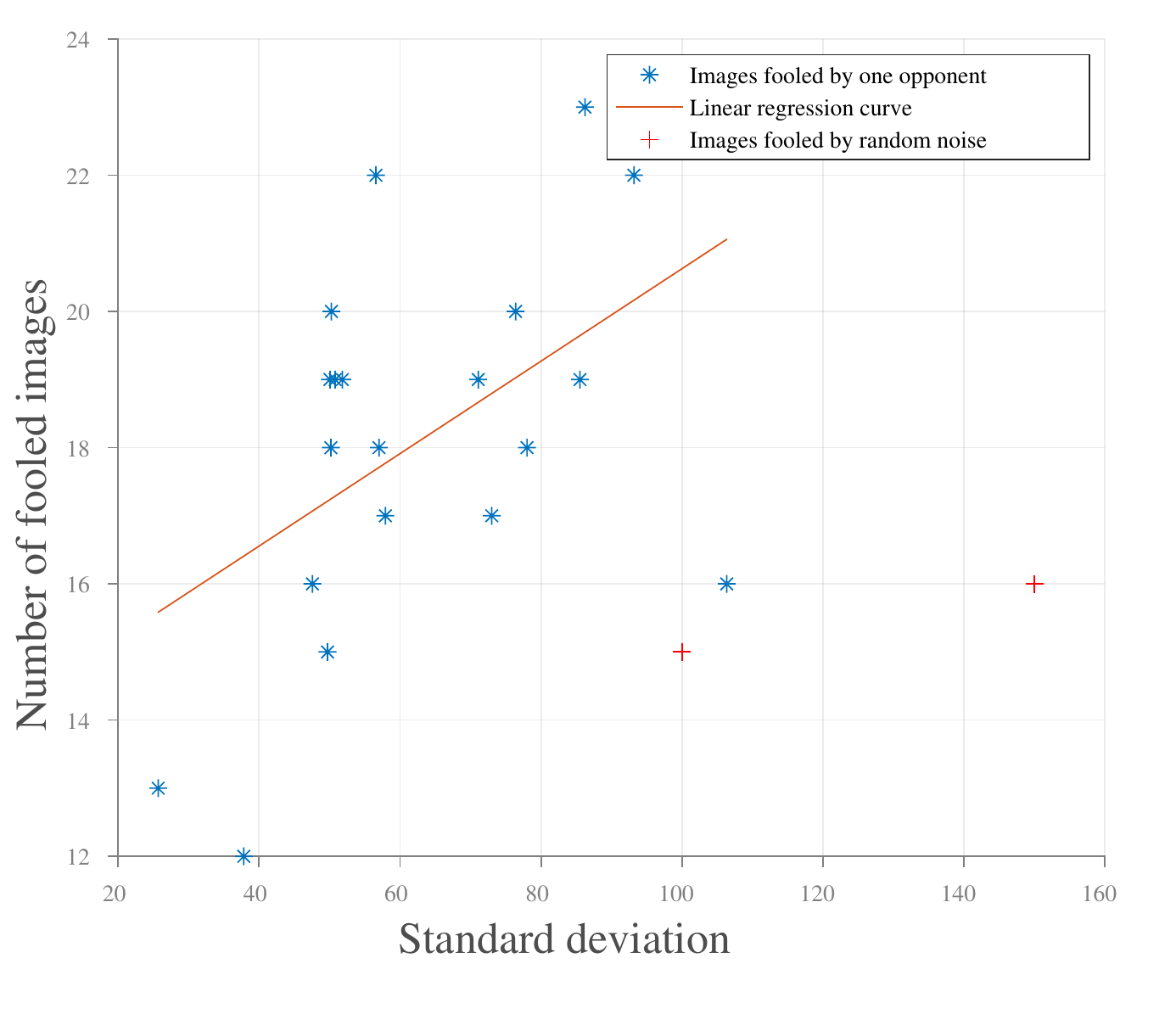}
                \caption{Total number of fooled images versus the standard deviations of the decoys. The transparency is equal to 4. Red cross are the gaussian noise.}
                \label{fig:stds}
            \end{figure}

\begin{figure}[t]
                \centering
                \includegraphics[width=0.8\linewidth,height=3.05cm]{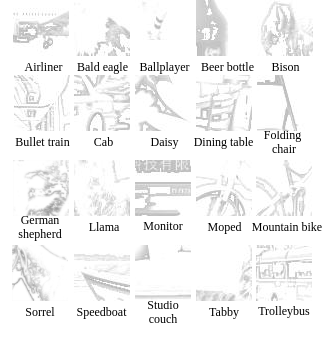}
                \caption{Decoys generated from the patches exposed in Figure \ref{fig:patches}}
                \label{fig:decoys}
           \end{figure}

        We report the results of Deepception on a subset of ImageNet consisting of 100 randomly selected images\footnote{Images were selected by randomly drawing ILSVRC2012 images (i.e. integers from [1, 100]), using the randperm function of the scientific computing environment Matlab after initializing Matlab random number generator seed with 0.}. We limit the experiment to 100 randomly selected images to obtain fair results that can be computed in a reasonable amount of time. This was inspired by recent approaches \cite{metzen2017detecting}.        
        We take the patches from figure \ref{fig:patches} and generate the decoys displayed in Figure \ref{fig:decoys}. These decoys are not perceptible by humans. We decided to observe the capacity of fooling of Deepception with 20 decoys (Figure \ref{fig:stds}). We observed a linear relation between standard deviation of the decoys and their fooling abilities. For this reason we designed 2 different gaussian noises (std=100 and 150). And we observed that a gaussian noise alone cannot provide the same fooling as our decoys.

We decided to apply our Deepception approach to a pre-trained VGG-VDD-19 DNN. The mountain-bike patch is chosen as a decoy because its exhibit high performance (high probability and high standard deviation).
 
        \subsubsection{Influence of different parameters of Deepception}

\begin{figure}[t]
            \subfloat[]{
                \includegraphics[width=0.5\linewidth]{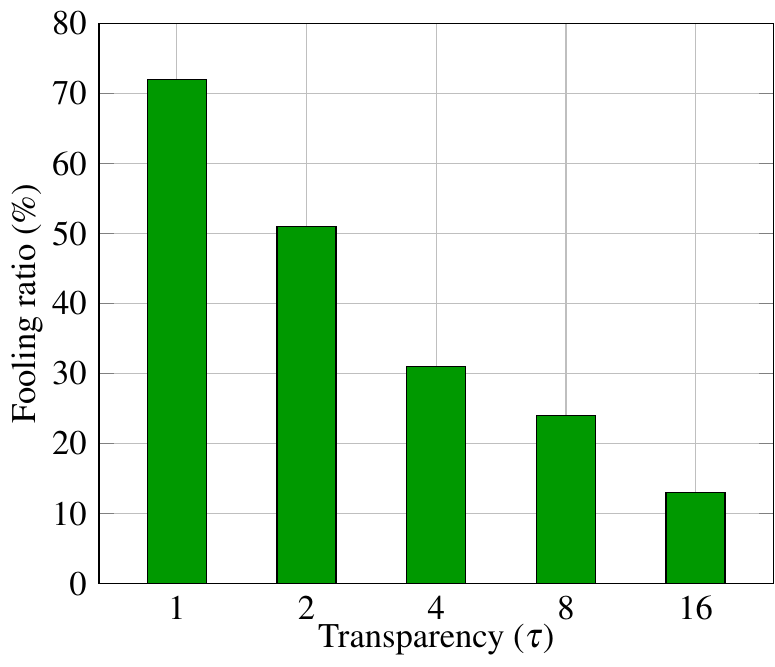}
                
                } 
                \subfloat[]{
                \includegraphics[width=0.5\linewidth]{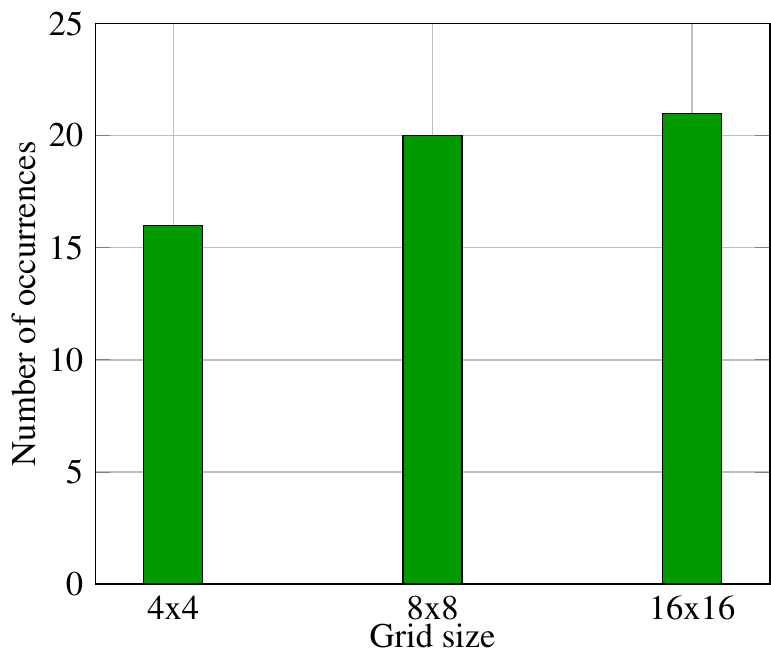}
                
                }
                \caption{(a) Fooling ratio on the validation set versus the decoy transparency. (b) Number of images fooled by positioning a single patch for different grid sizes.}
                \label{fig:transparency_sizes}
            \end{figure}

             Firstly, we fix the grid size to 4x4 and test 5 transparency levels to see how the fooling ratio varies. Figure \ref{fig:transparency_sizes}.a demonstrates the higher the transparency gets, the lower the fooling ratio will be.

            Transparency coefficient $\tau=4$ seems to be the optimal value for the 100 validation images. At this level of transparency, the decoy is unrecognizable to a human eye. Furthermore, the fooling ratio of 30\% for a 4x4 grid is acceptable. Consequently, we decide to fix $\tau=4$.
            
\begin{figure}
                \centering
                \subfloat[]{
                    \includegraphics[width=0.40\linewidth]{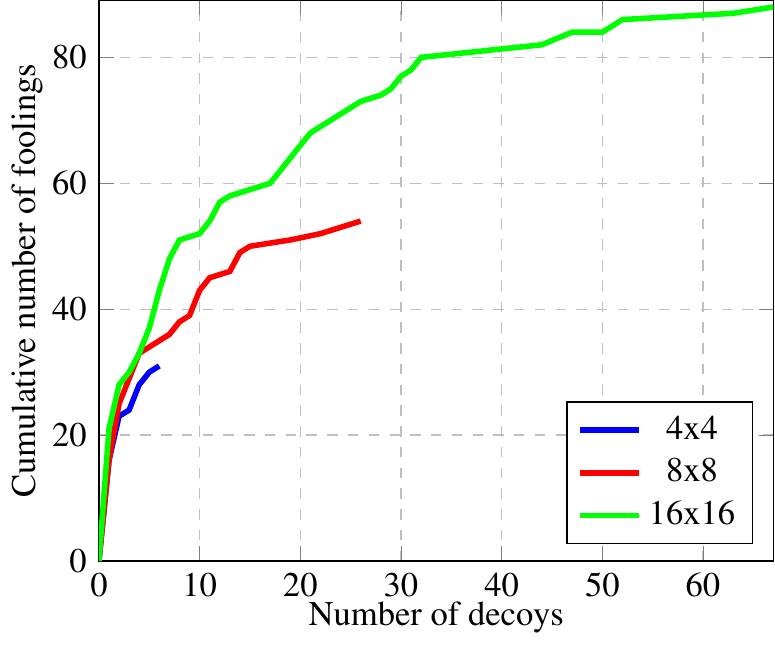}
                    }
                 
                 \subfloat[]{
                 \includegraphics[width=0.5\linewidth,height=3.0cm]{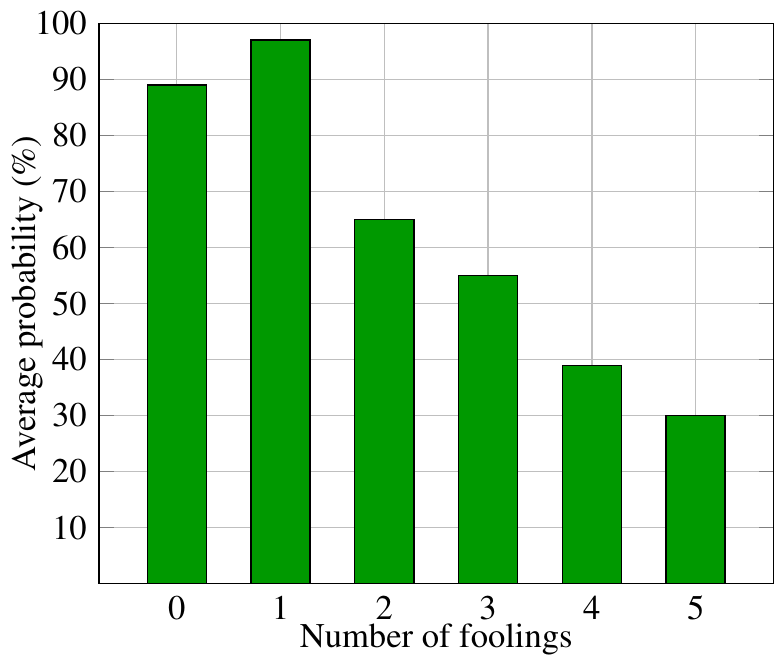}
                 }
                \caption{(a)Total number of fooled target images versus the number of inserted decoys. (b) Average of the targets' initial probabilities versus the number of foolings.  }
               \label{fig:nbFooled_nbDecoys}

            \end{figure}

            We want to study how many images are fooled with the first decoy placement. Figure \ref{fig:transparency_sizes}.b shows the number of images which have been fooled only by one decoy.
             Then, we investigate whether there is a relation between the initial probability of a target image and the number of times it gets fooled applying $\tau=4$ . Figure \ref{fig:nbFooled_nbDecoys}.b shows this relation actually exists between both variables.             
            Another important factor is how the number of decoys inserted inside the target, affects the performances. Figure \ref{fig:nbFooled_nbDecoys}.a reports these results for different grid sizes. The higher the number of inserted decoys gets, the higher the number of fooled image will be.
            
            \subsection{Comparison between Deepception and Universal  Adversarial Perturbation}
With a better understanding of the important parameters for Deepception, we decided to compare our results with a Universal Adversarial Perturbation \cite{moosavi2017universal}.        

\begin{figure}
            \centering
            \includegraphics[width=1.0\linewidth]{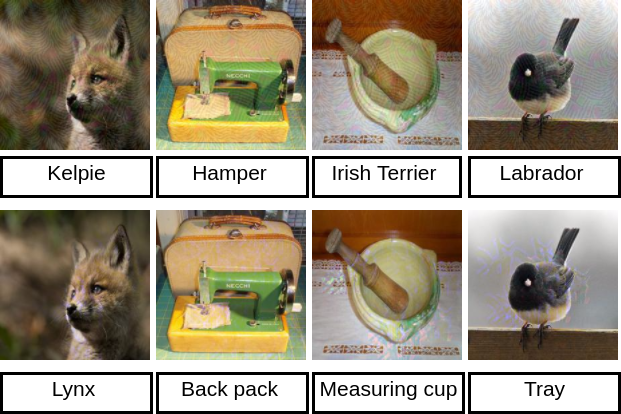}
            \caption{ Examples of perturbed images and the categories they belong to. Top row: Universal Adversarial Perturbation. Bottom row: our approach, Deepception. Zoom for a better visualization.}
            \label{fig:foolings}
        \end{figure}

       Some examples of fooled images by Deepception are illustrated in Figure \ref{fig:foolings}. The Top row illustrates Universal Adversarial Perturbation \cite{moosavi2017universal} and the Bottom row is Deepception, our approach. It can be seen that our approach is even less visible than the Universal Adversarial Perturbation.  
        
        We choose VGG-VDD-19 as the targeted deep neural network for the comparison. Therefore, we download from the Universal Adversarial Perturbations authors' online repository the pre-computed perturbation for this network ( 10,000 images of the ILSVRC 2012 \cite{ILSVRC15} training set). Afterwards, we randomly select 100 images from the ILSVRC 2012 validation dataset and apply Deepception and the Universal Perturbation.
        
        For the 100 randomly selected images, the fooling ratios of Deepception and Universal Adversarial Perturbations are 88\% and 75\% respectively. Figure \ref{fig:foolings} allows a visual comparison between the outputs of both methods. It is worthy noting that contrary to Universal Adversarial Perturbation, our approach does not need to be trained on a specific network.

\section{Discussion}
    This paper, formalized and analyzed some properties of DNNs. One of these properties can be considered as logical inferences of former studies. For instance,  it is possible to partially infer the local property, based on related studies on deconvolutional networks \cite{noh2015learning}, or visualization \cite{durand2017wildcat}, showing that some specific locations of an image respond better than others. But these works are mainly pixel based and did not study the effect of the patch size as we did.
 Other properties are more challenging to intuit. For instance, it is important to note that the spatial property reported here, is not related to an interaction between the patch and the content of the image because we show the effect in an image filled by zeros, excluding the patch. This property highlights that DNNs are not totally translation invariant. 
    
    However, the cumulative property allows to reinterpret the work of Nguyen et al. (2014) \cite{nguyen2015deep} and the reason why repeating a pattern from an object inside an image, increases its probability of being detected. Moreover, this paper exposes that repeating the pattern is not a sufficient condition to increase its probability of detection. In fact, it must be repeated at very specific locations of the image: the activating positions. Otherwise, the probability of detecting the object will not increase. Consequently, we highlighted a phenomenon that, at the best of our knowledge, no paper has reported yet: the activating-inhibitory property.
    The validation of these properties on Pascal VOC07 and ImageNet with different types of network proves that they can be generalized. 
    We also showed that contrary to a repetitive pattern, activation and inhibition are not mandatorily contiguous. The patch can benefit of the cumulative effect even with a sparse spatial distribution.
    
    This work also provides a new way of measuring the performance of a DNN: the ratio between minimum and maximum probability (for the same patch). The ratio should be equal to 1 to have a perfect translation-invariant networks. This measure can be optimized in the future to obtain more robust networks. We can observe that DNNs like Fast-RCNN-VGG16 or Resnet, which have proved to have state-of-the-art performances on the Pascal VOC07 and ImageNet datasets, are sensitive to patch translations. Their high ratio is meaning there is a big gap of probability between two different positions inside the image. 
    
    Some possible interpretation of the spatial property can be done based on "DNN neurophysiologist" studies of Bau et al. \cite{bau2017network} and Zhang et al. \cite{zhang2017interpretable}. The authors studied the activation of some high-level semantic units based on their receptive field. Our local property can be interpreted as the tendency of a given network to not be able to obtain many receptive fields that will cover uniformly the image. A conjoint framework combining psychophysics and neurophysiologist approach remains to be developed but has a great potential.
    
    The second part of this study took advantage of the exposed properties. We proposed a new fooling approach called Deepception. Unlike Universal Adversarial Perturbation (UAP), our approach uses structured patches from another class that are hidden in the image and are able to fool the network. We showed a fooling ratio of 88\% on the VGG-VDD-19 network compared to the 85\% obtained by UAP on the same network \cite{moosavi2017universal}.        

   An fundamental difference of our approach compared to classical approaches is that we do not need to have a prior knowledge of the network's architecture. We simply need to be able to send an input and an access to the probability. Another difference is also the sparsity. In many cases, we can fool the network only with 1 patch (it represents $1/16$ th of the image). The advantage is that very localized patch could be more difficult to perceive by a human than a widespread fooling noise. We also tested that our effect is stronger with a structured patch rather than using a simple Gaussian noise. And even if we show that variance impacts the capacity of fooling the target, a Gaussian noise with high variance is not better for fooling compared with a decoy selected with our technique. We observed also that the same decoy can be utilized to fool different images. Our approach exhibits a good generality. Indeed, when a decoy is selected with a given network, it can be employed to fool many other images of the same network.

\section{Conclusion}
In this article we used a "Psychophysic" approach applied to Artificial Intelligence in the realm of "Neurophysiologists". We did not study the internal architectures of the networks, but we made some deductions by modifying the input and analyzing the resulting probability. We think this approach can have some benefit and we propose some properties allowing to rank the networks and explaining their high performances. With a practical application, a software called Deepception, we demonstrated that the properties analyzed in this work can help designing new methods for deep neural networks' fooling.


{\small
\bibliographystyle{named}
\bibliography{egbib}
}

\end{document}